\documentclass[runningheads]{llncs}
\usepackage[T1]{fontenc}
\usepackage{graphicx,verbatim}
\usepackage{amssymb}
\usepackage{bbding}
\usepackage{pifont}
\newcommand{\mycross}{\ding{55}} 
\usepackage{booktabs}
\usepackage{dsfont}
\usepackage{algorithm}
\usepackage{algorithmic}
\usepackage{amsmath}
\usepackage{multirow}

\usepackage[misc]{ifsym}

\usepackage{fontawesome}

\begin{document}
%
\title{MCA-RG: Enhancing LLMs with Medical Concept Alignment for Radiology Report Generation}
%

\author{Qilong Xing \and Zikai Song\textsuperscript{$^{\textrm{\Letter}}$} \and 
Youjia Zhang \and Na Feng \and \\ Junqing Yu \and Wei Yang\textsuperscript{$^{\textrm{\Letter}}$}}  
\authorrunning{Q. Xing et al.}
\institute{Huazhong University of Science and Technology, Wuhan, China \\
\email{\{qlxing, skyesong, weiyangcs\}@hust.edu.cn}}

\maketitle              

\let\thefootnote\relax\footnotetext{${\textrm{\Letter}}$ Corresponding authors: skyesong@hust.edu.cn, weiyangcs@hust.edu.cn}
\begin{abstract}
Despite significant advancements in adapting Large Language Models (LLMs) for radiology report generation (RRG), clinical adoption remains challenging due to difficulties in accurately mapping pathological and anatomical features to their corresponding text descriptions. Additionally, semantic agnostic feature extraction further hampers the generation of accurate diagnostic reports. 
To address these challenges, we introduce Medical Concept Aligned Radiology Report Generation (MCA-RG), a knowledge-driven framework that explicitly aligns visual features with distinct medical concepts to enhance the report generation process. MCA-RG utilizes two curated concept banks: a pathology bank containing lesion-related knowledge, and an anatomy bank with anatomical descriptions. 
The visual features are aligned with these medical concepts and undergo tailored enhancement. We further propose an anatomy-based contrastive learning procedure to improve the generalization of anatomical features, coupled with a matching loss for pathological features to prioritize clinically relevant regions. Additionally, a feature gating mechanism is employed to filter out low-quality concept features. 
Finally, the visual features are corresponding to individual medical concepts, and are leveraged to guide the report generation process. 
Experiments on two public benchmarks (MIMIC-CXR and CheXpert Plus) demonstrate that MCA-RG achieves superior performance, highlighting its effectiveness in radiology report generation.


\end{abstract}

\section{Introduction}

Radiography remains the primary imaging modality for diagnosing thoracic conditions (e.g., pneumonia and lung cancer detection via chest X-rays) due to its cost-effectiveness and widespread availability~\cite{mimic_cxr_database}. However, manual analysis of radiographs and subsequent report generation impose significant cognitive burdens on radiologists. 
Automatic radiology report generation (RRG) has emerged as a critical tool to alleviate radiologists’ workloads while enhancing diagnostic consistency by synthesizing clinically precise and structured reports from medical images.
The development for RRG has evolved from early CNN-RNN frameworks \cite{jing2020show} to more advanced Transformer-based approaches \cite{warm_start,m2tr_pro}. Notable advancements include the integration of memory-driven mechanisms \cite{chen2022cross,r2gen}, contrastive learning techniques \cite{liu2021contrastive}, and the incorporation of knowledge graphs \cite{kiut}.
Recently, Large Language Models (LLMs) have demonstrated remarkable success in text generation tasks and have been increasingly applied to radiology report generation \cite{minigpt,AdaMatchCyclic,orid,meddr,xraygpt}. However, they still struggle to produce ``clinically reliable reports'' that align with radiologists’ diagnostic reasoning.

A primary challenge is the accurate mapping of pathological and anatomical features to their corresponding descriptions. Existing methods that rely on conflated image features for report generation may suffer from feature interference, leading to incorrect mappings, such as erroneously generating pathological findings from anatomical features or vice versa. 
Furthermore, extracting discriminative visual features for pathologies and anatomies remains challenging: anatomical variability across patients \cite{roth2015anatomy} introduces heterogeneity in structural patterns, while pathological features suffer from spatial sparsity and contextual noise \cite{mitigating,VLCI}, as lesions often occupy localized regions within radiographs.

Common remedies involve leveraging detected abnormal anatomical regions to guide report generation, thereby enhancing both clinical relevance and model interpretability \cite{rgrg}. However, these approaches heavily rely on {high-quality anatomical bounding box annotations}, which are scarce and typically restricted to a narrow subset of anatomical structures \cite{imagenome}. 

To address these challenges, we propose \textbf{M}edical \textbf{C}oncept \textbf{A}ligned Radiology \textbf{R}eport \textbf{G}eneration (MCA-RG), 
a method that utilizing visual features corresponding to distinct medical concepts to guide the report generation process. 
We first extract medical concepts from reports, organizing them into anatomy and pathology banks, and enriching their semantic representations using publicly available LLMs. Image features are aligned with these concepts and integrated with relevant medical knowledge. 
To enhance visual features, we propose an anatomy-based contrastive learning method to improve the generalization of anatomical features and introduce an matching loss to encourage the model to focus on pathology-related regions. Additionally, we propose a concept feature gating mechanism to filter out low-quality visual features. 
Finally, instead of using conflated features like existing methods, we guide the LLM with enhanced concept-specific visual features. 
Embedded medical knowledge clarifies the association between visual features and medical concepts in reports, while precise visual features further refine the accuracy of the generated descriptions.

\begin{figure*}[t]
  \centering
    \includegraphics[width=1\linewidth]{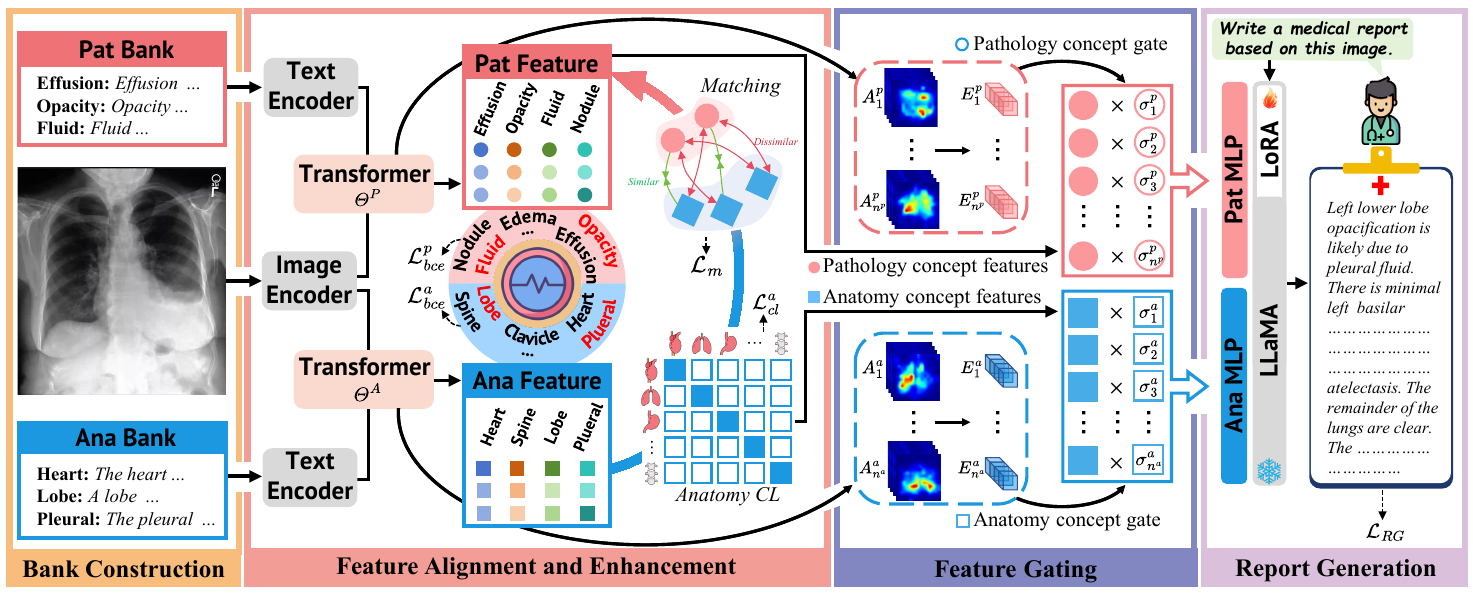}
    \caption{
Our MCA-RG first extracts and enriches the pathological and anatomical concepts with medical knowledge, and then align the image features with concepts by guiding the model to diagnose pathologies and anatomical abnormalities. Anatomical features are enhanced through anatomy-based contrastive learning, while pathological features are refined using matching loss. Finally, a gating mechanism filters the visual features of each concept before they are used for medical report generation.
    }
    \label{fig:overall_arch}
  \hfill
\end{figure*}

\section{Method}
Our MCA-RG framework consists of four stages, as illustrated in Fig. \ref{fig:overall_arch}.

\subsection{Pathology and Anatomy Bank Construction}
In medical reports, clinicians document pathological findings alongside the anatomical regions where they are located. To exploit this, we propose constructing concept banks to pre-store these clinical elements.
Specifically, we apply a Named Entity Recognition (NER) method, RadGraph \cite{radgraph}, to identify entities related to anatomies and pathologies within the report and use them as medical concepts. RadGraph models the relationship between anatomies and pathologies by representing them as \{pathology, anatomy, exist\} triplets, for instance, [opacity, lung, True]. 
Consequently, given a medical report $\mathcal{T}$, we transform the sentences into triplets of the form:
\begin{equation}
    \mathcal{T} = \{C_i^p, C_j^a, E_{ij}\}_{n^p \times n^a}, i \in [1, n^p], j \in [1, n^a],
\end{equation}
Here, $E_{ij} \in \{0,1\}$ indicates whether the pathology $C_i^p$ is present at the anatomy $C_j^a$. Additionally, $n^p$ and $n^a$ represent the number of concepts in the pathology and anatomy banks, respectively. 
To enrich each medical concept, we enhance them with additional medical knowledge using GPT-4 for detailed explanations.

\subsection{Feature Alignment and Enhancement}
We propose a feature alignment and enhancement process to extract precise, concept-specific visual features embedded with medical knowledge. 
Our approach first uses a pre-trained language model \cite{clinical_bert} to convert banked concepts into text embeddings, while encoding the radiology image $\mathcal{I}$ with a ResNet-50 visual encoder:
\begin{equation}
    \mathcal{V} = \Phi _{visual}(\mathcal{I}), \quad 
    \{t_i^p\}_{n^p}, \{t_j^a\}_{n^a} = \Phi _{textual}(\{C_i^p\}_{n^p}, \{C_j^a\}_{n^a}), 
\end{equation}

\textbf{Anatomical Features.}
To align features with anatomical concepts, we propose training the model to diagnose whether an anatomical structure is healthy, defined by the absence of any notable pathologies. 
The alignment process is accomplished by a Transformer module, denoted as $\Theta^A$, which comprises multiple Transformer Decoder layers \cite{detr}. Each anatomical concept embedding, $t_j^a$, is used as the \textit{Query}, while the visual features are utilized as the \textit{Key} and \textit{Value}. To guide this procedure, the aligned visual feature of each anatomical concept are further transformed using a linear projection matrix $W^a \in \mathbb{R}^{d \times 2}$, and the anatomical-level labels $Y^a$ indicating anatomy healthy status are employed for supervision: 
\begin{equation}
    v_{j}^{'',a} = \Theta^A(t_{j}^a W_{q}^a, \mathcal{V} W_{k}^a, \mathcal{V} W_{v}^a ),\quad \mathcal{L}_{bce}^a = \frac{1}{n^a}\sum_{j=1}^{n^a} \mathcal{L}_{bce}( v_{j}^{'',a} W^a, Y_j^a),
\end{equation}
where $W_{q}$, $W_{k}$, and $W_{v}$ are projection matrices used in the attention mechanism. Besides the alignment procedure, the module also fuses the concept embeddings with the aligned features to enhance them with medical knowledge.

To address anatomical variability, we propose an anatomy-based contrastive learning method. 
Specifically, an MLP module is employed to transform the extracted anatomical features, $\{v_j^{'',a}\}_{n^a}$, into a more robust representation, $\{v_j^{',a}\}_{n^a}$. 
This transformation is optimized using contrastive learning, which is applied to anatomical features from different samples. Positive pairs are defined as features representing the same anatomical structure, and negative pairs are defined as features representing different anatomical structures. The learning process is guided by minimizing the contrastive loss, $\mathcal{L}_{cl}^a$: 
\begin{equation}
    \mathcal{L}_{cl}^a = -\frac{1}{n^a}\sum_{j=1}^{n^a} \mathrm{log} \frac{\mathrm{exp(sim}(v_{j}^{',a}, v_{j,k_0}^{',a}))}{\sum_{j'=1}^{n^a}\mathrm{exp(sim}(v_{j}^{',a}, v_{j',k_0,}^{',a}))},
\end{equation}
where $\{v_{j,k_0}^{',a}\}_{n^a}$ represents enhanced anatomical features from another sample and $\mathrm{sim(,)}$ refers to cosine similarity. By minimizing the loss term $\mathcal{L}_{cl}^a$, the model is encouraged to identify general anatomical features shared across samples of the same anatomy, such as anatomical locations, while retaining the ability to distinguish between different anatomical structures. To generate anatomical features for input to the LLM, we further employ an MLP module to fuse $\{v_j^{'',a}\}_{n^a}$ and $\{v_j^{',a}\}_{n^a}$, resulting in enhanced anatomical features $\{v_j^{a}\}_{n^a}$.

\textbf{Pathological Features.}
Similar to the procedure of anatomical feature alignment, we guide the model to align pathological features $\{v_{i}^p\}_{n^p}$ by training it with pathology-level labels $Y^p$ and loss $\mathcal{L}_{bce}^p$ to diagnose specific pathologies.

Considering the model may be affected by features from irrelevant areas, we propose a pathology-anatomy matching loss. Since enhanced anatomical features capture how abnormalities manifest within specific structures, we increase the feature similarity between pathologies and anatomies containing these pathologies, while reducing the similarity between pathologies and anatomies without them. The loss is only applied to present pathologies:

\begin{equation}
    \mathcal{L}_{m} = \frac{1}{\mathrm{sum}(Y^p)}\sum_{i=1}^{n^p} Y_i^p \times \frac{1}{n^a} \sum_{j=1}^{n^a} |E_{ij} - \mathrm{max[sim}(v_{i}^p,{v_{j}^{a})}, 0]|,
\end{equation}
The loss guides the model to focus on disease-related regions while being less affected by features from unrelated areas, thereby improving diagnostic accuracy.

\subsection{Feature Gating}

To ensure the LLM effectively uses informative visual features, we propose a Feature Gating (FG) mechanism to suppresses low-quality concept features, reducing noise and ensuring that meaningful visual features enhance the report generation process. 
Specifically, the quality of feature extraction is evaluated using attention maps derived from the concept feature alignment modules ($\Theta^A$ and $\Theta^P$). We argue that attention maps with a more uniform value distribution indicate less effective feature extraction, as the model may struggle to concentrate on critical regions. 
To quantify this, we compute the entropy of the attention map for each concept and each head within the multi-head attention mechanism.

Since attention maps for different concepts may vary in distribution, a high absolute entropy does not always indicate low-quality visual tokens. To account for concept-specific variations and aggregate information across attention heads, we apply a linear projection for each concept to map multi-head entropy to a 
single gating value, which is then used for feature filtering:
\begin{equation}
E_i^p = \mathrm{Entropy}(A_i^p)\in \mathbb R^{\text{head\_num}}, \;
    \sigma_i^p = \mathrm{Sigmoid}(E_i^p W_i^{g,p}), \; v_{i}^{g,p} = \sigma_i^p \times v_{i}^{p},
\end{equation}
where $A_i^p$ denotes the multi-head attention maps and $W_i^{g,p} \in \mathbb{R}^{\text{head\_num} \times 1}$ represents the projection matrix for the entropy values of the $i$-th pathology concept. The procedure for anatomical concepts is similar and omitted for brevity.

\subsection{Report Generation}

Unlike existing methods that input conflated image features into LLMs, we use features aligned with medical concepts to leverage the embedded medical knowledge and precise visual information, enabling accurate mapping between visual features and textual descriptions. To align the visual encoder's feature space with the LLM, we introduce separate MLP modules to project pathological and anatomical features. The report generation is optimized by minimizing cross-entropy loss with ground truth sequences $\{y_t\}_T$:
\begin{equation}
\mathcal{L}_{RG} = -\sum_{t=1}^{T} \mathrm{log} p(y_t|y_1,...,y_{t-1},\\  \mathrm{MLP^P}(\{v_{i}^{g,p}\}_{n^p}),\mathrm{MLP^A}(\{v_{j}^{g,a}\}_{n^a})),
\end{equation}
In summary, the training objective combines multiple loss functions:
\begin{equation}
    \mathcal{L} = \beta_0 \times (\mathcal{L}_{bce}^a + \mathcal{L}_{bce}^p) + \beta_1 \times (\mathcal{L}_{m} + \mathcal{L}_{cl}^a) + \mathcal{L}_{RG}.
\end{equation}

\section{Experiments}

\subsection{Dataset and Implementation Details}
\textbf{Dataset.} 
We conduct experiments on MIMIC-CXR \cite{mimic_cxr_database} and CheXpert Plus \cite{chexpertplus}, following the preprocessing method used in R2Gen \cite{r2gen}. MIMIC-CXR uses predefined data splits \cite{warm_start}, while CheXpert Plus is split into 52,505/1,500/3,800 for training/validation/test after filtering out reports without a findings section.

\textbf{Evaluation Metrics.}
We employ both Natural Language Generation (NLG) metrics and Clinical Efficacy (CE) metrics for evaluation, following prior works \cite{r2gen,warm_start}.
For NLG evaluation, we compute BLEU-n \cite{bleu}, ROUGE-L \cite{rouge} and METEOR  \cite{meteor} score. 
To evaluate clinical accuracy, we employ clinical metrics as outlined in WarmStart \cite{warm_start}.

\textbf{Implementation Details.}
Our MCA-RG fine-tunes LLaMA2 (7B) \cite{llama2} using LoRA modules, with both rank and alpha set to 64. 
During the inference stage for report generation, the beam size for beam search is set to 3. 
Medical concepts are selected based on their frequency in MIMIC-CXR reports, resulting in 51 anatomical and 67 pathological concepts. These are also used for CheXpert, as the most frequent concepts in CheXpert are covered by those in MIMIC-CXR. 
By default, we set $\beta _0 = 0.5$ and $\beta _1 = 0.3$ to balance the different loss terms.

\begin{table*}[t]
\centering
\caption{Comparison of our method with prior works in NLG scores (BLEU (BL), METEOR (MTR), ROUGE-L (RG-L)). Best results are in bold, second-best underlined. Scores are cited from original papers and \cite{r2llm,hergen}. For LLM-based methods, the number of parameters is provided. * indicates methods re-implemented by us following official code.
}
\renewcommand\arraystretch{1}
\scalebox{1}{
\begin{tabular}{llcccccccc}
\toprule[1.1pt]
Dataset                     & Method    &Year      & LLM  & BL-1         & BL-2         & BL-3         & BL-4         & MTR            & RG-L           \\ \midrule
\multirow{9}{*}{MIMIC-CXR} & R2Gen\cite{r2gen}   & 2020        &\mycross      & 0.353          & 0.218          & 0.145          & 0.103          & 0.142          & 0.277          \\
                            & R2GenCMN\cite{chen2022cross}   & 2022     &\mycross & 0.353          & 0.218          & 0.148          & 0.106          & 0.142          & 0.278          \\
                            & WarmStart\cite{warm_start}  & 2023     &\mycross & \underline{0.392}          & \underline{0.245}          & \underline{0.169}          & \underline{0.124}          & 0.153          & 0.285          \\
                            & XrayGPT\cite{xraygpt}  & 2023       & 7B   & 0.128          & 0.045          & 0.014          & 0.004          & 0.079          & 0.111          \\
                            & ORID\cite{orid}   & 2024         & 7B   & 0.386          & 0.238          & 0.163          & 0.117          & 0.15           & 0.284          \\
                            & MedDr\cite{meddr}   & 2024       & 34B  & 0.322          & -              & -              & 0.072          & \textbf{0.238} & 0.226          \\
                            & AMC\cite{AdaMatchCyclic} & 2024 & 3B   & 0.379          & 0.235          & 0.154          & 0.106          & \underline{0.163}          & \underline{0.286}          \\ 
                            & MiniGPT-Med*\cite{minigpt} & 2024 & 7B   & 0.336          & 0.201          & 0.130          & 0.091          & 0.128          & 0.250         \\ \cline{2-10} 
                            & Ours    & -        & 7B   & \textbf{0.411} & \textbf{0.256} & \textbf{0.175} & \textbf{0.128} & \underline{0.163}          & \textbf{0.300} \\
                            \cline{1-10}
                            \multirow{5}{*}{CheXpertPlus} 
                             & R2Gen*\cite{r2gen} & 2020 & \mycross   & 0.325          & 0.195          & 0.130          & 0.090          & 0.129          & 0.262         \\
                              & R2GenCMN*\cite{chen2022cross} & 2022 & \mycross   & 0.336          & 0.200          & 0.131          & 0.090          & 0.133          & \underline{0.264}         \\
                               & WarmStart*\cite{warm_start} & 2023 & \mycross   & \underline{0.350}          & \underline{0.207}          & \underline{0.134}          & \underline{0.091}          & \underline{0.134}          & 0.263         \\
                                & MiniGPT-Med*\cite{minigpt} & 2024 & 7B   & 0.341          & 0.194          & 0.121          & 0.079          & 0.125          & 0.255         \\
                                \cline{2-10}
                                 & Ours & - & 7B   & \textbf{0.367}          & \textbf{0.218}          & \textbf{0.149}          & \textbf{0.102}          & \textbf{0.147}          & \textbf{0.266}         \\
                            \bottomrule[1.1pt]
\end{tabular}
}
\label{tab:nlg_table}
\end{table*}

\subsection{Main Results}

We compare MCA-RG with competitive RRG methods, demonstrating superior performance in both NLG and CE metrics. For NLG (Table \ref{tab:nlg_table}), it achieves the highest ROUGE-L scores of 0.3 and 0.266 on two benchmarks. In clinical accuracy (Table \ref{tab:ce_table}), it aciveves both highest score on example-based F1 score and macro F1 score. 
Among LLM-based methods, MCA-RG substantially outperforms models with the same parameter count, further highlighting its effectiveness.

\begin{table*}[t]
\centering
\caption{Comparison of our method with prior works on the MIMIC-CXR test set in CE scores (Precision (P), Recall (R), F1). Best results are in bold, second-best underlined. * indicates our re-implementation following official code. Other scores and CE computation methods (example-based or macro) are cited from \cite{warm_start,hergen} and official GitHub sources.}
\renewcommand\arraystretch{1.0}
\scalebox{1}{
\begin{tabular}{clcccccccc}
\toprule[1.1pt]
\multicolumn{1}{l}{\multirow{2}{*}{Dataset}}   & \multicolumn{1}{l}{\multirow{2}{*}{Method}}  & \multicolumn{1}{l}{\multirow{2}{*}{Year}} & \multicolumn{1}{l}{\multirow{2}{*}{LLM}} & \multicolumn{3}{c|}{Example-based} & \multicolumn{3}{c}{Macro-based}   \\ \cline{5-10} 
                           & \multicolumn{1}{c}{}  & \multicolumn{1}{c}{} & \multicolumn{1}{c}{} & P    & R   & \multicolumn{1}{c|}{F1}      & P & R & F1    \\ \midrule
\multirow{6}{*}{MIMIC-CXR} 
                           & R2Gen\cite{r2gen}  & 2020 &\mycross & 0.333         & 0.273    & \multicolumn{1}{c|}{0.276}   & 0.297      & 0.189  & 0.193 \\
                           & R2GenCMN\cite{chen2022cross} &2022 &\mycross & 0.334         & 0.275    & \multicolumn{1}{c|}{0.278}   & 0.354      & 0.271  & 0.275 \\
                           & WarmStart\cite{warm_start} &2023 &\mycross & 0.418         & 0.367    & \multicolumn{1}{c|}{\underline{0.367}}   & \underline{0.417}      & \underline{0.295}  & \underline{0.306} \\
                           & ORID\cite{orid} & 2024 & 7B & \underline{0.435}         & 0.295    & \multicolumn{1}{c|}{0.352}   & -      & -  & - \\
                           & MiniGPT-Med*\cite{minigpt} & 2024 & 7B & 0.335         & 0.245    & \multicolumn{1}{c|}{0.264}   & 0.236      & 0.188  & 0.175 \\
                           \cline{2-10} 
                           & Ours & - & 7B & \textbf{0.472}         & \underline{0.406}    & \multicolumn{1}{c|}{\textbf{0.408}}   & \textbf{0.443}      & \textbf{0.306}  & \textbf{0.335} \\ \bottomrule[1.1pt]
\end{tabular}
}
\label{tab:ce_table}
\end{table*}

\begin{figure*}[t]
  \centering
    \includegraphics[width=1\linewidth]{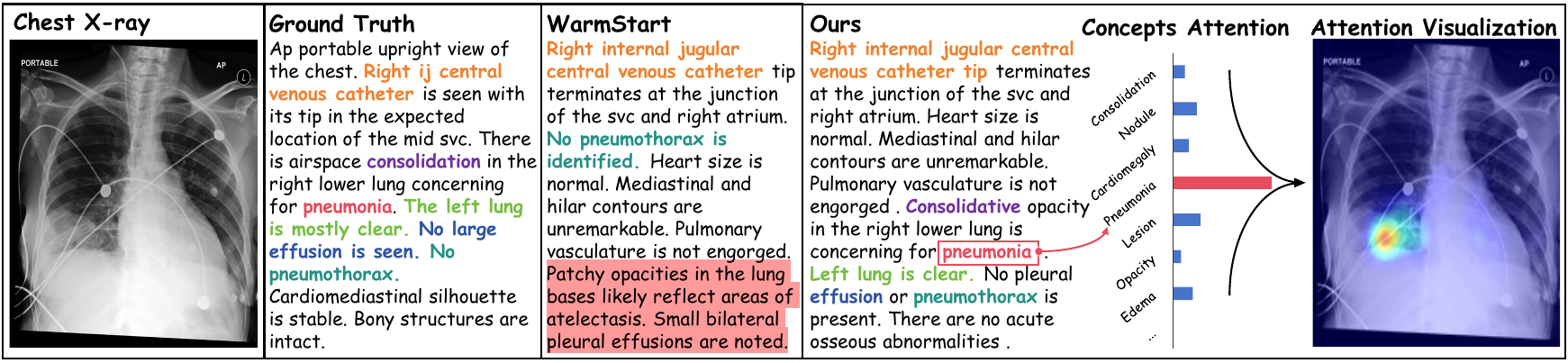}
    \caption{
    An illustration comparing the reports generated by different models for a given input radiograph. Each matched descriptions are highlighted using consistent colors to facilitate comparison.
    Red background indicates wrong descriptions. 
    }
    \label{fig:report_vis}
  \hfill
\end{figure*}

\begin{figure}[t]
\centering
\begin{minipage}[t]{0.48\linewidth}
\centering
\includegraphics[width=0.8\linewidth]{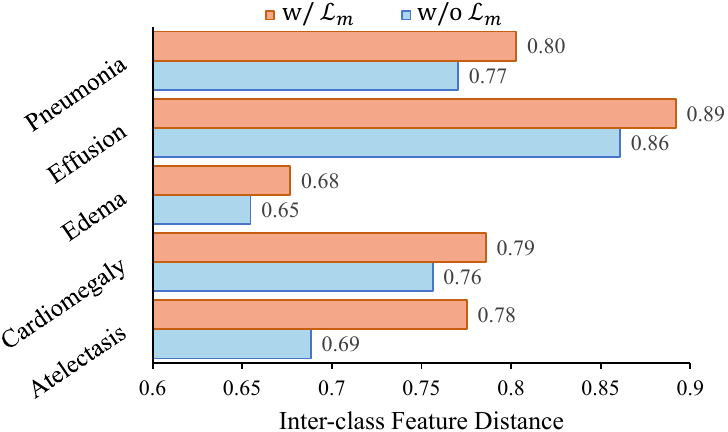}
\caption{The visualization of inter class feature distance.}
\label{fig:chexpert_feat_dis}
\end{minipage}
\begin{minipage}[t]{0.48\linewidth}
\centering
\includegraphics[width=0.8\linewidth]{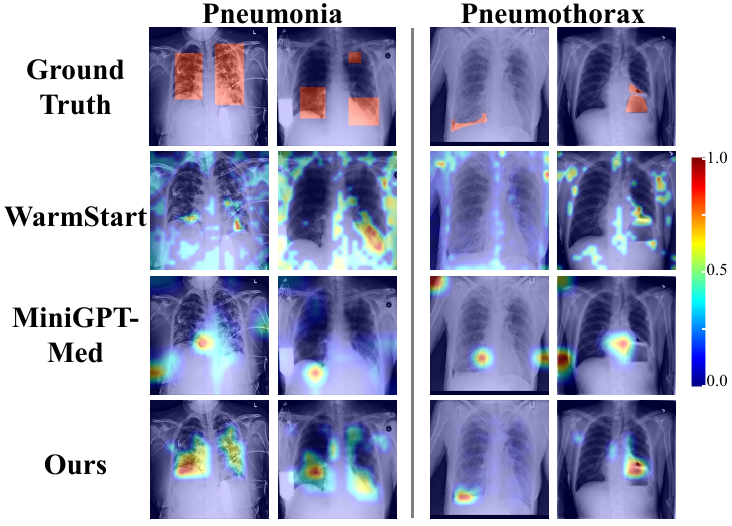}
\caption{The visualization of attention map corresponding to term ”pneumonia” (left) and ”pneumothorax” (right)}
\label{fig:obs_grounding}
\end{minipage}
\end{figure}

\begin{table*}[t]
\centering
\caption{The ablation study of different operations in terms of NLG metrics (BLEU (BL), METEOR (MTR), ROUGE-L (RG-L), average improvement across all NLG metrics over the baseline (AVG.$\Delta$)) and macro F1 on MIMIC-CXR dataset. 
}
\scalebox{1}{
\begin{tabular}{cllllccccccccccc}
\toprule[1.1pt]
$\mathcal{L}_{bce}^p$ & $\mathcal{L}_{bce}^a$ & $\mathcal{L}_{cl}^a$ & $\mathcal{L}_{m}$ & FG & BL-1 & BL-2 & BL-3 & BL-4 & MTR   & RG-L  & \multicolumn{1}{c|}{AVG.$\Delta$} & $\rm F_{1}$  \\ \midrule
&      &  &  &           & 0.332  & 0.209  & 0.143  & 0.104  & 0.131 & 0.280 & \multicolumn{1}{c|}{-}  &0.205    \\
\checkmark & \checkmark &  & &  & 0.382  & 0.237  & 0.158  & 0.112  & 0.152 & 0.283 & \multicolumn{1}{c|}{+10.6\%}  &0.288 \\
\checkmark & \checkmark & \checkmark & &  & 0.397  & 0.246  & 0.164  & 0.117  & 0.157 & 0.284 & \multicolumn{1}{c|}{+14.3\%}  &0.291 \\
\checkmark & \checkmark & & \checkmark & & 0.396  & 0.248  & 0.165  & 0.116  & 0.156 & 0.285 & \multicolumn{1}{c|}{+14.2\%}  &0.284 \\
\checkmark & \checkmark & \checkmark & \checkmark & & 0.403  & 0.252  & 0.173  & 0.126  & \textbf{0.163} & 0.289 & \multicolumn{1}{c|}{+18.6\%}  &0.324 \\
\checkmark & \checkmark & \checkmark & \checkmark & \checkmark & \textbf{0.411}  & \textbf{0.256}  & \textbf{0.175}  & \textbf{0.128}  & \textbf{0.163} & \textbf{0.300} & \multicolumn{1}{c|}{+20.5\%}  & \textbf{0.335} \\
                            \bottomrule[1.1pt]
\end{tabular}
}
\label{tab:ablation_nlg}
\end{table*}

\subsection{Model Analysis}

\textbf{Case Study.} Fig.~\ref{fig:report_vis} shows an example of a generated report. While WarmStart misses \textit{pneumonia} and produces false positives like \textit{atelectasis} and \textit{effusion}, our method accurately identifies \textit{pneumonia} and its location (\textit{right lower lung}) while correctly diagnosing the absence of \textit{effusion}, significantly enhancing clinical accuracy.
To understand this performance, we analyze the attention weights from the concepts bank and visualize the attention map. 
The visualization illustrates that when generating the term \textit{pneumonia}, there is increased attention on the corresponding pathological concept, while the concept features effectively capture useful information from disease-related regions.

\textbf{Ablation Study.}
Ablation experimental results are presented in Table \ref{tab:ablation_nlg}. 
The feature alignment process, guided by the loss terms $\mathcal{L}_{bce}^p$ and $\mathcal{L}_{bce}^a$, achieves a 10.6\% ($\Delta+$10.6\%) improvement in NLG metrics and raises the macro F1 score to 0.288. 
Applying feature enhancement independently for different features further improves NLG performance, while combining both methods yields an 8\% ($\Delta+$8\%) boost in NLG metrics and increases the macro F1 score from 0.288 to 0.324. The FG mechanism ensures the LLM effectively utilizes concept visual features, leading to additional performance gains. 
To analyze the effect of matching loss, we create a $5 \times 200$ MIMIC-CXR dataset following MedCLIP \cite{medclip}. This dataset includes samples from five classes: Atelectasis, Cardiomegaly, Edema, Effusion, and Pneumonia. 
For each class, we compute the centroids and calculate the average inter-class conceptual feature distances using cosine similarity. 
Our analysis reveals that the matching loss increases the model's ability to distinguish between different classes (see Fig.\ref{fig:chexpert_feat_dis}).

\textbf{Attention Visualization.} 
To evaluate the model's ability to focus on disease-related regions, we extract attention maps during predictions of medical terms (e.g., pneumonia) to identify the image regions referenced. Using the RSNA \cite{rsna} and SIIM-ACR Pneumothorax \cite{siimacr} datasets with fine-grained disease annotations (see Fig.\ref{fig:obs_grounding}), 
we find that WarmStart and MiniGPT-Med often focus on unrelated regions, whereas our method effectively targets key areas.

\section{Conclusion}

In this work, we propose MCA-RG for RRG task by leveraging the visual features of medical concepts. Embedded medical knowledge enhances the mapping from visual features to text, while feature enhancement refines representations for clinically accurate descriptions. 
In the future, we aim to evolve our approach into an interactive GPT-like system for question answering.

\bibliographystyle{splncs04}
\bibliography{references}

\begin{thebibliography}{10}
\providecommand{\url}[1]{\texttt{#1}}
\providecommand{\urlprefix}{URL }
\providecommand{\doi}[1]{https://doi.org/#1}

\bibitem{mitigating}
Ahmadi, R., Rajabi, M.J., Khalooie, M., Sabokrou, M.: Mitigating bias: Enhancing image classification by improving model explanations. arXiv preprint arXiv:2307.01473  (2023)

\bibitem{minigpt}
Alkhaldi, A., Alnajim, R., Alabdullatef, L., Alyahya, R., Chen, J., Zhu, D., Alsinan, A., Elhoseiny, M.: Minigpt-med: Large language model as a general interface for radiology diagnosis. arXiv preprint arXiv:2407.04106  (2024)

\bibitem{clinical_bert}
Alsentzer, E., Murphy, J.R., Boag, W., Weng, W.H., Jin, D., Naumann, T., McDermott, M.: Publicly available clinical bert embeddings. arXiv preprint arXiv:1904.03323  (2019)

\bibitem{meteor}
Banerjee, S., Lavie, A.: Meteor: An automatic metric for mt evaluation with improved correlation with human judgments. In: Proceedings of the acl workshop on intrinsic and extrinsic evaluation measures for machine translation and/or summarization. pp. 65--72 (2005)

\bibitem{detr}
Carion, N., Massa, F., Synnaeve, G., Usunier, N., Kirillov, A., Zagoruyko, S.: End-to-end object detection with transformers. In: European conference on computer vision. pp. 213--229. Springer (2020)

\bibitem{chexpertplus}
Chambon, P., Delbrouck, J.B., Sounack, T., Huang, S.C., Chen, Z., Varma, M., Truong, S.Q., Chuong, C.T., Langlotz, C.P.: Chexpert plus: Augmenting a large chest x-ray dataset with text radiology reports, patient demographics and additional image formats (2024), \url{https://arxiv.org/abs/2405.19538}

\bibitem{VLCI}
Chen, W., Liu, Y., Wang, C., Zhu, J., Zhao, S., Li, G., Liu, C.L., Lin, L.: Cross-modal causal intervention for medical report generation. arXiv preprint arXiv:2303.09117  (2023)

\bibitem{AdaMatchCyclic}
Chen, W., Shen, L., Lin, J., Luo, J., Li, X., Yuan, Y.: Fine-grained image-text alignment in medical imaging enables explainable cyclic image-report generation. In: Proceedings of the 62nd Annual Meeting of the Association for Computational Linguistics (Volume 1: Long Papers). pp. 9494--9509 (2024)

\bibitem{chen2022cross}
Chen, Z., Shen, Y., Song, Y., Wan, X.: Cross-modal memory networks for radiology report generation. arXiv preprint arXiv:2204.13258  (2022)

\bibitem{r2gen}
Chen, Z., Song, Y., Chang, T.H., Wan, X.: Generating radiology reports via memory-driven transformer. arXiv preprint arXiv:2010.16056  (2020)

\bibitem{orid}
Gu, T., Yang, K., An, X., Feng, Z., Liu, D., Cai, W.: Orid: Organ-regional information driven framework for radiology report generation. arXiv preprint arXiv:2411.13025  (2024)

\bibitem{meddr}
He, S., Nie, Y., Chen, Z., Cai, Z., Wang, H., Yang, S., Chen, H.: Meddr: Diagnosis-guided bootstrapping for large-scale medical vision-language learning. arXiv preprint arXiv:2404.15127  (2024)

\bibitem{kiut}
Huang, Z., Zhang, X., Zhang, S.: Kiut: Knowledge-injected u-transformer for radiology report generation. In: Proceedings of the IEEE/CVF Conference on Computer Vision and Pattern Recognition. pp. 19809--19818 (2023)

\bibitem{radgraph}
Jain, S., Agrawal, A., Saporta, A., Truong, S.Q., Duong, D.N., Bui, T., Chambon, P., Zhang, Y., Lungren, M.P., Ng, A.Y., et~al.: Radgraph: Extracting clinical entities and relations from radiology reports. arXiv preprint arXiv:2106.14463  (2021)

\bibitem{jing2020show}
Jing, B., Wang, Z., Xing, E.: Show, describe and conclude: On exploiting the structure information of chest x-ray reports. arXiv preprint arXiv:2004.12274  (2020)

\bibitem{mimic_cxr_database}
Johnson, A.E., Pollard, T.J., Berkowitz, S.J., Greenbaum, N.R., Lungren, M.P., Deng, C.y., Mark, R.G., Horng, S.: Mimic-cxr, a de-identified publicly available database of chest radiographs with free-text reports. Scientific data  \textbf{6}(1), ~317 (2019)

\bibitem{rouge}
Lin, C.Y.: Rouge: A package for automatic evaluation of summaries. In: Text summarization branches out. pp. 74--81 (2004)

\bibitem{r2llm}
Liu, C., Tian, Y., Chen, W., Song, Y., Zhang, Y.: Bootstrapping large language models for radiology report generation. In: Proceedings of the AAAI Conference on Artificial Intelligence. vol.~38, pp. 18635--18643 (2024)

\bibitem{liu2021contrastive}
Liu, F., Yin, C., Wu, X., Ge, S., Zou, Y., Zhang, P., Sun, X.: Contrastive attention for automatic chest x-ray report generation. arXiv preprint arXiv:2106.06965  (2021)

\bibitem{warm_start}
Nicolson, A., Dowling, J., Koopman, B.: Improving chest x-ray report generation by leveraging warm starting. Artificial intelligence in medicine  \textbf{144},  102633 (2023)

\bibitem{m2tr_pro}
Nooralahzadeh, F., Gonzalez, N.P., Frauenfelder, T., Fujimoto, K., Krauthammer, M.: Progressive transformer-based generation of radiology reports. arXiv preprint arXiv:2102.09777  (2021)

\bibitem{bleu}
Papineni, K., Roukos, S., Ward, T., Zhu, W.J.: Bleu: a method for automatic evaluation of machine translation. In: Proceedings of the 40th annual meeting of the Association for Computational Linguistics. pp. 311--318 (2002)

\bibitem{roth2015anatomy}
Roth, H.R., Lee, C.T., Shin, H.C., Seff, A., Kim, L., Yao, J., Lu, L., Summers, R.M.: Anatomy-specific classification of medical images using deep convolutional nets. In: 2015 IEEE 12th international symposium on biomedical imaging (ISBI). pp. 101--104. IEEE (2015)

\bibitem{rsna}
Shih, G., Wu, C.C., Halabi, S.S., Kohli, M.D., Prevedello, L.M., Cook, T.S., Sharma, A., Amorosa, J.K., Arteaga, V., Galperin-Aizenberg, M., et~al.: Augmenting the national institutes of health chest radiograph dataset with expert annotations of possible pneumonia. Radiology: Artificial Intelligence  \textbf{1}(1),  e180041 (2019)

\bibitem{rgrg}
Tanida, T., M{\"u}ller, P., Kaissis, G., Rueckert, D.: Interactive and explainable region-guided radiology report generation. In: Proceedings of the IEEE/CVF Conference on Computer Vision and Pattern Recognition. pp. 7433--7442 (2023)

\bibitem{xraygpt}
Thawkar, O., Shaker, A., Mullappilly, S.S., Cholakkal, H., Anwer, R.M., Khan, S., Laaksonen, J., Khan, F.S.: Xraygpt: Chest radiographs summarization using medical vision-language models. arXiv preprint arXiv:2306.07971  (2023)

\bibitem{llama2}
Touvron, H., Martin, L., Stone, K., Albert, P., Almahairi, A., Babaei, Y., Bashlykov, N., Batra, S., Bhargava, P., Bhosale, S., et~al.: Llama 2: Open foundation and fine-tuned chat models. arXiv preprint arXiv:2307.09288  (2023)

\bibitem{hergen}
Wang, F., Du, S., Yu, L.: Hergen: Elevating radiology report generation with longitudinal data. arXiv preprint arXiv:2407.15158  (2024)

\bibitem{medclip}
Wang, Z., Wu, Z., Agarwal, D., Sun, J.: Medclip: Contrastive learning from unpaired medical images and text. arXiv preprint arXiv:2210.10163  (2022)

\bibitem{imagenome}
Wu, J.T., Agu, N.N., Lourentzou, I., Sharma, A., Paguio, J.A., Yao, J.S., Dee, E.C., Mitchell, W., Kashyap, S., Giovannini, A., et~al.: Chest imagenome dataset for clinical reasoning. arXiv preprint arXiv:2108.00316  (2021)

\bibitem{siimacr}
Zawacki, A., Wu, C., Shih, G., Elliott, J., Fomitchev, M., Hussain, M., Lakhani, P., Culliton, P., Bao, S.: Siim-acr pneumothorax segmentation (2019)

\end{thebibliography}

\end{document}